\documentclass{article}

\usepackage{arxiv}

\usepackage[utf8]{inputenc} 
\usepackage[T1]{fontenc}    
\usepackage{hyperref}       
\usepackage{url}            
\usepackage{booktabs}       
\usepackage{amsfonts}       
\usepackage{nicefrac}       
\usepackage{microtype}      
\usepackage{lipsum}
\usepackage{graphicx}
\graphicspath{ {./images/} }
 \usepackage{amsmath}
\usepackage{algorithm}%
\usepackage{algorithmicx}%
\usepackage{algpseudocode}%
\usepackage{lipsum}
\usepackage{graphicx}
\usepackage{doi}

\usepackage{arxiv}

\usepackage[utf8]{inputenc} 
\usepackage[T1]{fontenc}    
\usepackage{hyperref}       
\usepackage{url}            
\usepackage{booktabs}       
\usepackage{amsfonts}       
\usepackage{nicefrac}       
\usepackage{microtype}      
\usepackage{lipsum}		
\usepackage{graphicx}

\usepackage{doi}

\title{MT-Depth: Multi-task Instance feature analysis for the Depth Completion}

\author{
Abdul Haseeb Nizamani \\
Saisuode (Shanghai) Intelligent Technology Co., Ltd. (Synthoid.ai)\\
Shanghai, China \\
\texttt{ni.jianma@synthoid.ai} \\
\and
Dandi Zhou \\
Saisuode (Shanghai) Intelligent Technology Co., Ltd. (Synthoid.ai)\\
Shanghai, China \\
\texttt{zhou.dandi@synthoid.ai} \\
\and
Xinhai Sun\\
Saisuode (Shanghai) Intelligent Technology Co., Ltd. (Synthoid.ai)\\
Shanghai, China \\
\texttt{sun.xinhai@synthoid.ai} \\
}

\date{}
\renewcommand{\shorttitle}{MT-Depth: Multi-task Instance Feature Analysis for the Depth Completion}

\begin{document}
\maketitle
\begin{abstract}

Depth completion plays a vital role in 3D perception systems, especially in scenarios where sparse depth data must be densified for tasks such as autonomous driving, robotics, and augmented reality. While many existing approaches rely on semantic segmentation to guide depth completion, they often overlook the benefits of object-level understanding. In this work, we introduce an instance-aware depth completion framework that explicitly integrates binary instance masks as spatial priors to refine depth predictions. Our model combines four main components: a frozen YOLO V11 instance segmentation branch, a U-Net-based depth completion backbone, a cross-attention fusion module, and an attention-guided prediction head. The instance segmentation branch generates per-image foreground masks that guide the depth branch via cross-attention, allowing the network to focus on object-centric regions during refinement. We validate our method on the Virtual KITTI 2 dataset, showing that it achieves lower Root Mean Squared Error (RMSE) compared to both a U-Net-only baseline and previous semantic-guided methods, while maintaining competitive Mean Absolute Error (MAE). Qualitative and quantitative results demonstrate that the proposed model effectively enhances depth accuracy near object boundaries, occlusions, and thin structures. Our findings suggest that incorporating instance-aware cues offers a promising direction for improving depth completion without relying on dense semantic labels.

\end{abstract}


\section{Introduction}

Many CNN-based depth completion models adopt an encoder–decoder architecture that takes RGB and sparse depth as input, using skip connections to preserve spatial details \cite{klingner2020selfsupervised, Eldesokey2020confidence, Li2024SemanticGuidedDC}. Despite their success, single-task models often struggle in challenging scenarios—such as thin structures, distant objects, and occlusion boundaries—where geometric cues alone are insufficient. To overcome these challenges, recent research has explored the integration of semantic information to guide depth estimation \cite{AtapourAbarghouei2019ToCO, Huang2019indoor}. Semantic segmentation introduces class-level context, enabling the network to recognize object boundaries and propagate meaningful structure across regions of missing data.

Multi-task learning (MTL) has emerged as a natural extension, allowing models to jointly learn depth, segmentation, and other geometric cues within a shared framework \cite{ZHAO2023106496, Moeskops2016, Shivakumar2019, Cheng2019CSPNLC}. By leveraging shared representations across related tasks, MTL not only improves generalization but also facilitates information transfer between domains. For instance, predicting surface normals alongside depth can enhance 3D consistency, while semantic segmentation helps identify meaningful object boundaries that align with depth discontinuities.

Among these, SemSegDepth \cite{Lagos_2022} is a notable example that jointly predicts dense depth and semantic segmentation maps from RGB and sparse depth inputs. It uses a shared encoder with separate decoder branches for each task and demonstrates that fusing semantic features improves depth quality. Similarly, PanDepth \cite{Lagos2022PanDepthJP} extends the idea to panoptic segmentation, integrating semantic, instance, and depth understanding into a single multi-task model.

Despite the effectiveness of semantic cues, most prior work has focused on pixel-level class labels, overlooking the potential of instance-level features. Instance segmentation—where each object instance is labeled separately—offers a richer and more structured representation of the scene. Instance masks encode spatial extents, shapes, and object-specific boundaries, making them particularly useful for refining depth in occluded or cluttered regions. However, integrating instance-aware reasoning into depth completion has received limited attention.

In real-world applications like robotics and autonomous driving, instance-level understanding is critical. Accurate depth prediction around objects—especially for small, distant, or partially occluded entities—is essential for collision avoidance, manipulation, and interaction. Therefore, combining instance segmentation with depth completion offers the potential to significantly enhance 3D scene understanding \cite{Lee2021LearningMD, Jo2024, s24072374}.

Recent advances in real-time instance segmentation, particularly the YOLO family, provide efficient and accurate object detection and mask prediction \cite{Bochkovskiy2020yolov4, khanam2024yolov5deeplookinternal, YOLOv8}. YOLO V11, a recent version, introduces instance mask heads alongside object detection outputs, enabling fast extraction of binary masks at multiple scales \cite{Khanam2024YOLOv11AO}. These instance-aware features are well-suited for integration with depth completion, as they provide per-object guidance without requiring pixel-level semantic class labels.

In this work, we propose a novel multi-task architecture that explicitly leverages instance-level features to improve depth completion. Our model consists of four main components: (1) a frozen instance segmentation branch based on YOLO V11, (2) a U-Net-based depth completion branch, (3) a cross-attention module that fuses instance and depth features, and (4) an attention-guided prediction head that outputs the final dense depth map. By treating instance masks as attention queries, the model is encouraged to focus on object regions and align depth predictions with structural boundaries.

The segmentation branch uses YOLO V11 to generate per-image binary instance masks, which are merged into a foreground object map. These masks act as spatial priors that highlight regions of interest. The depth branch receives a concatenation of RGB, sparse depth, and a binary validity mask. The U-Net processes this input and outputs both initial depth predictions and multi-scale feature maps.

To fuse the two modalities, we employ a cross-attention mechanism where instance features act as queries and depth features as keys and values. This guides the network to focus on relevant object regions. The resulting attention-enhanced features are fused with the original depth features using a feature fusion module. Finally, an attention-based head refines the fused representation and produces the dense depth output.

We validate our model on the Virtual KITTI 2 dataset \cite{cabon2020vkitti2, Gaidon2016vkitti}, a synthetic benchmark offering perfect ground truth for depth, instance segmentation, and RGB under varied lighting and weather conditions. Our results show that the proposed instance-aware architecture outperforms baseline depth completion models and achieves competitive accuracy compared to semantic-guided approaches—without requiring class-level labels. Our model reduces RMSE while maintaining low MAE, especially in regions near object edges and thin structures.

Our contributions are summarized as follows:

\begin{itemize}
  \item We introduce a depth completion framework that explicitly integrates instance segmentation via cross-attention, using binary masks from a frozen YOLO V11 model.
  \item We show that instance-level guidance improves depth predictions in cluttered, occluded, and boundary regions.
  \item We demonstrate state-of-the-art performance on Virtual KITTI 2, achieving better depth quality compared to baselines that do not exploit instance-aware features.
\end{itemize}

Our findings highlight the importance of instance-aware priors for depth completion and open new directions for combining detection and 3D perception in real-time systems. Future work may explore extending this approach to real datasets, optimizing for low-power deployment, or incorporating full panoptic understanding for richer scene modeling.

\section{Related Works}

\subsection{Instance Segmentation}

Recent progress in instance segmentation has accelerated thanks to advances in architectures, loss functions, and data-efficient learning strategies. A 2025 survey reviews over 200 deep‑learning based segmentation models, categorizing them by backbone, mask decoding, and temporal consistency strategies — a useful resource to track current trends and limitations of mask-based segmentation methods. \cite{XIA2025128740, tomography11050052} At the same time, novel frameworks like diffusion‑based instance segmentation have emerged: the authors of a 2025 study demonstrate that diffusion models can generate high‑quality instance masks even for challenging cases such as overlapping objects or weak boundaries. \cite{Ma2025diffusion} For transparent or translucent objects — traditionally hard to segment — a recent few‑shot method leverages data augmentation and template consistency to reliably produce instance masks from limited examples. \cite{Cherian2024TrInSeg} On the application side, instance segmentation remains fundamental for tasks such as robot bin picking, scene understanding, and autonomous perception. \cite{Xu2025VIS} Finally, contemporary surveys of segmentation methods highlight the trade‑offs among speed, mask fidelity, and generalization. \cite{Wolk2024survey}

These advances confirm that instance segmentation today is not only more accurate, but also more robust and efficient. In our work, we build on this progress by using a frozen instance segmentation model to provide object‑level masks for depth completion, benefiting from the recent improvements in mask quality and model generalization.

\subsection{Depth Completion}

Depth completion remains a hot topic as it bridges sparse sensor outputs (e.g. LiDAR or partial depth) with dense depth estimation required for perception or 3D reconstruction. A newly proposed method from late 2024, SigNet, treats depth completion as a two‑step problem: first densify sparse depth using non‑CNN tools, then enhance the coarse depth via a CNN. This “degradation-aware” paradigm helps alleviate artifacts caused by direct CNN completion from very sparse inputs. \cite{Yan2024SigNet} Meanwhile, hybrid depth‑guided sensor fusion approaches are also gaining traction, this approach describes a model that combines LiDAR and camera data with semantic and geometric fusion to achieve robust perception under challenging conditions. \cite{Tim2025} On the 3D completion side, a recent method integrates stereo depth cues and implicit depth-aware features to complete full 3D occupancy and semantic maps — a promising direction for applications like autonomous driving and scene reconstruction. \cite{Miao2023OccDepth} Additionally, transformer-based and multi‑path aggregation networks have been developed for semantic scene completion, demonstrating strong performance on large-scale driving datasets. \cite{Xia2023SCPNet} Finally, a 2023 survey of depth and scene reconstruction methods outlines major challenges and categorizes modern approaches, providing useful context for evaluating new models. \cite{Rajapaksha2024}

Our approach differs from these by explicitly incorporating instance‑level semantic priors into the depth completion pipeline, rather than relying solely on geometric cues or generic RGB guidance. This enables better depth prediction around object boundaries and in occluded regions.

\subsection{Multi‑Task Learning}

Joint learning of multiple related tasks — such as depth estimation, segmentation, surface normals, or scene reconstruction — has shown consistent benefits by enabling cross-task information sharing and improved generalization. A 2025 work proposes a novel multi-task framework with a shared encoder and separate decoders for surface normals and depth, demonstrating improved consistency across depth and surface predictions \cite{CHAVARRIASSOLANO2025103379}. Meanwhile, the transformer-based framework SwinMTL (2024) performs concurrent depth estimation and semantic segmentation on monocular images, using adversarial training to refine predictions and showing strong performance across both tasks \cite{Taghavi2024SwinMTL}. Combining depth completion and object detection has also been explored in \cite{Pan2024ForegroundAware}, this work introduces a foreground‑aware network that fuses depth and object detection tasks using a transformer-based architecture, improving both depth completion and detection under sparse data conditions.  In addition, generative‑adversarial methods have been proposed to jointly learn segmentation and depth from monocular images, showing that adversarial signals can improve the realism and consistency of depth maps. \cite{Zhang2021MTGAN} Finally, a broader survey of multi‑task dense prediction methods published in 2024 outlines key challenges in balancing losses, architecture design, and cross-task interference. \cite{li2024multitask}

Our method adopts this multi-task philosophy, but with a twist: rather than learning instance segmentation jointly, we use a pretrained segmentation model (frozen) and focus on depth completion — using instance masks as attention priors. This design reduces training complexity while leveraging the strength of instance-level cues, and avoids issues related to joint backpropagation across unrelated tasks.

\section{Method}

\subsection{Segmentation Branch Using YOLO V11}

YOLO (You Only Look Once) is a real-time object detection framework that simultaneously performs object classification and localization in a single pass through a neural network \cite{Khanam2024YOLOv11AO}. It divides the input image into a fixed grid and predicts bounding boxes, objectness scores, and class probabilities directly for each cell. This one-stage design contrasts with earlier two-stage detectors like R-CNN, making YOLO significantly faster while maintaining strong accuracy. Over time, the YOLO family has evolved through various iterations (v1 to v10), each improving on backbone architecture, feature fusion, and multi-scale detection performance.

YOLO V11 introduces several enhancements, including an improved backbone with deeper layers and more efficient upsampling for better spatial resolution in segmentation tasks. Most notably, it incorporates an instance segmentation head that outputs per-object binary masks, allowing the network to perform pixel-wise segmentation in addition to bounding box detection. These masks are aligned with bounding boxes using a learned mask coefficient, producing a full-resolution instance prediction. The segmentation head enables YOLO V11 to support applications beyond object detection, such as panoptic segmentation and instance-aware understanding, all in real-time.

In our model, we adopt a pretrained YOLO V11 instance segmentation model to form the segmentation branch. This branch remains frozen throughout training and is used to generate binary instance masks for each RGB input image. Given an RGB image $\mathbf{I} \in \mathbb{R}^{3 \times H \times W}$, the YOLO model $\mathcal{Y}$ returns a set of predicted masks $\{\mathbf{M}_1, \dots, \mathbf{M}_n\}$, where each $\mathbf{M}_i \in \{0,1\}^{H_i \times W_i}$ represents an instance mask resized to the resolution of detection. To obtain a unified binary foreground mask, we perform an element-wise maximum operation across all instance masks:

\begin{equation}
\mathbf{M}_{\text{merged}} = \max_{i=1}^{n} \mathbf{M}_i,
\end{equation}

which yields a single-channel mask $\mathbf{M}_{\text{merged}} \in [0,1]^{H \times W}$ indicating all foreground object regions. This merged mask is resized to match the feature resolution used in the attention module and treated as a low-dimensional semantic prior. To match the expected input format, the mask is reshaped to include the batch and channel dimensions:

\begin{equation}
\mathbf{M}_{\text{seg}} = \text{Interpolate}(\mathbf{M}_{\text{merged}}, \text{size}=(H', W')) \in \mathbb{R}^{B \times 1 \times H' \times W'}.
\end{equation}

This feature map is used as the query input in a cross-attention mechanism, where depth features serve as keys and values. By doing so, we guide the network to pay more attention to object-centric regions during the depth refinement process. Notably, the YOLO model is used in a frozen state, reducing computational cost and preserving the reliability of its pretrained segmentation capabilities. This fusion of semantic segmentation with geometric depth information improves the model's focus and accuracy, especially around object boundaries and occlusion edges.

\subsection{Depth Completion Branch Using U-Net}

The depth completion branch is responsible for estimating dense depth maps from sparse depth inputs and RGB images \cite{Ronneberger2015, Wang2022, Ramesh2023SIUNetSI}. This is achieved using a modified U-Net architecture, which takes as input a concatenation of the RGB image, sparse depth, and a binary mask indicating valid depth pixels. Let $\mathbf{I} \in \mathbb{R}^{3 \times H \times W}$ denote the RGB image, $\mathbf{D}_{\text{sparse}} \in \mathbb{R}^{1 \times H \times W}$ the sparse depth map, and $\mathbf{V} \in \{0,1\}^{1 \times H \times W}$ the corresponding validity mask. The combined input to the U-Net is:

\begin{equation}
\mathbf{X}_{\text{depth}} = \text{Concat}(\mathbf{I}, \mathbf{D}_{\text{sparse}}, \mathbf{V}) \in \mathbb{R}^{5 \times H \times W}.
\end{equation}

The U-Net encoder progressively downsamples the input through convolution and pooling layers, extracting multi-scale features while preserving local context. These features are then passed through a bottleneck and upsampled in the decoder using transposed convolutions, with skip connections from the encoder to preserve spatial detail.

The network outputs two components: a dense feature map $\mathbf{F}_{\text{depth}} \in \mathbb{R}^{C \times H \times W}$ and an initial dense depth prediction $\hat{\mathbf{D}}_{\text{init}} \in \mathbb{R}^{1 \times H \times W}$. The final layer of the U-Net applies a $1 \times 1$ convolution to the decoder output to generate the depth estimate:

\begin{equation}
\hat{\mathbf{D}}_{\text{init}} = \text{Conv}_{1\times1}(\mathbf{F}_{\text{depth}}).
\end{equation}

This initial depth serves both as a standalone prediction and as a foundation for further refinement through the fusion with semantic features. By including the validity mask as input, the model learns to differentiate between observed and unobserved regions, reducing overfitting to sparse data and improving generalization.

The U-Net’s skip connections and shared spatial resolution between input and output make it especially effective for dense prediction tasks like depth completion, where fine details and global structure both matter. This branch thus provides a geometrically grounded initial estimate that is further refined through the fusion with semantic segmentation features in the attention head.

\subsection{Cross-Attention}

To effectively fuse semantic information from instance masks with geometric depth features, we introduce a cross-attention module that allows the model to selectively focus on object regions during depth refinement. The core idea is to treat features extracted from the segmentation branch as queries and features from the depth backbone as keys and values, enabling the model to learn attention maps that emphasize semantically relevant areas.

Let $\mathbf{F}_{\text{seg}} \in \mathbb{R}^{C_s \times H \times W}$ denote the instance mask feature map (e.g., a binary map with $C_s = 1$), and let $\mathbf{F}_{\text{depth}} \in \mathbb{R}^{C_d \times H \times W}$ represent the depth feature map from the U-Net backbone. We first project the input features into a shared attention space using $1 \times 1$ convolutions:

\begin{equation}
\mathbf{Q} = \text{Conv}_{1\times1}^q(\mathbf{F}_{\text{seg}}), \quad
\mathbf{K} = \text{Conv}_{1\times1}^k(\mathbf{F}_{\text{depth}}), \quad
\mathbf{V} = \text{Conv}_{1\times1}^v(\mathbf{F}_{\text{depth}}),
\end{equation}

where $\mathbf{Q}, \mathbf{K}, \mathbf{V} \in \mathbb{R}^{C_a \times H \times W}$ and $C_a$ is the attention embedding dimension.

We reshape these tensors to apply scaled dot-product attention in the spatial domain. Let $N = H \times W$ be the number of spatial locations. The tensors are reshaped as follows:

\begin{equation}
\mathbf{Q}', \mathbf{K}', \mathbf{V}' \in \mathbb{R}^{N \times C_a}, \quad \text{where} \quad \mathbf{Q}' = \text{reshape}(\mathbf{Q}), \quad \mathbf{K}' = \text{reshape}(\mathbf{K}), \quad \mathbf{V}' = \text{reshape}(\mathbf{V}).
\end{equation}

The attention weights are computed using the scaled dot-product between queries and keys:

\begin{equation}
\mathbf{A} = \text{Softmax}\left( \frac{\mathbf{Q}' \cdot {\mathbf{K}'}^\top}{\sqrt{C_a}} \right) \in \mathbb{R}^{N \times N}.
\end{equation}

The attended features are then obtained by multiplying the attention matrix with the value tensor:

\begin{equation}
\mathbf{O}' = \mathbf{A} \cdot \mathbf{V}' \in \mathbb{R}^{N \times C_a}.
\end{equation}

Finally, we reshape $\mathbf{O}'$ back to the spatial form and apply a $1 \times 1$ convolution followed by a nonlinearity:

\begin{equation}
\mathbf{F}_{\text{att}} = \text{ReLU}(\text{Conv}_{1\times1}^{\text{out}}(\text{reshape}^{-1}(\mathbf{O}'))) \in \mathbb{R}^{C_d \times H \times W}.
\end{equation}

This attention-enhanced feature map $\mathbf{F}_{\text{att}}$ captures object-aware context that is semantically informed by the instance segmentation mask. It is later fused with the original depth features to improve depth estimation, particularly in foreground regions where accurate structure is critical.


\subsection{Prediction Head}

The prediction head refines the fused depth features and outputs the final dense depth map. This component is designed to focus selectively on informative channels through a lightweight attention mechanism, followed by a shallow convolutional decoder that maps high-dimensional fused features to a single-channel depth output.

Given the fused feature map $\mathbf{F}_{\text{fused}} \in \mathbb{R}^{C \times H \times W}$ produced by the feature fusion module, the prediction head first applies a channel-wise attention mechanism to emphasize important channels. This is achieved using a squeeze-and-excitation block inspired by SENet, implemented via global average pooling followed by two fully connected layers and a sigmoid activation:

\begin{equation}
\mathbf{z} = \frac{1}{H \cdot W} \sum_{i=1}^{H} \sum_{j=1}^{W} \mathbf{F}_{\text{fused}}(:,i,j), \quad \mathbf{s} = \sigma(W_2 \cdot \text{ReLU}(W_1 \cdot \mathbf{z})),
\end{equation}

where $\mathbf{z} \in \mathbb{R}^{C}$ is the channel-wise descriptor, $W_1 \in \mathbb{R}^{C/r \times C}$ and $W_2 \in \mathbb{R}^{C \times C/r}$ are learnable weights of the attention MLP with reduction ratio $r$, and $\mathbf{s} \in [0,1]^C$ are the attention weights. These weights are broadcast and applied to the feature map as:

\begin{equation}
\mathbf{F}_{\text{att}}(c, i, j) = \mathbf{s}(c) \cdot \mathbf{F}_{\text{fused}}(c, i, j).
\end{equation}

Next, the attention-weighted features are passed through two convolutional layers with ReLU activation and batch normalization to regress the final depth map:

\begin{equation}
\hat{\mathbf{D}}_{\text{final}} = \text{Conv}_{3\times3}^2\left( \text{ReLU} \left( \text{BN} \left( \text{Conv}_{3\times3}^1(\mathbf{F}_{\text{att}}) \right) \right) \right) \in \mathbb{R}^{1 \times H \times W}.
\end{equation}

Here, $\text{Conv}_{3\times3}^1$ maps $C$ channels to a mid-level dimension (e.g., 64), and $\text{Conv}_{3\times3}^2$ maps this to a single-channel output. The prediction head thus converts the fused semantic-geometric representation into a refined and fully dense depth map that benefits from both the structural context of the U-Net and the object-aware cues from the segmentation stream.

This design helps the network focus more strongly on semantically meaningful regions (e.g., vehicles, buildings) during final prediction, particularly in ambiguous areas where sparse depth measurements are missing or unreliable.

\subsection{Overall Model Architecture}

Our proposed framework for depth completion combines semantic and geometric information using a modular architecture consisting of four core components: a frozen YOLO V11 instance segmentation branch, a U-Net-based depth completion backbone, a cross-attention module, and an attention-guided prediction head.

The segmentation branch uses a pretrained YOLO V11 model to extract instance masks from RGB images. These masks are merged into a single-channel foreground map representing all object regions. Importantly, the YOLO model remains frozen during training, acting solely as a feature provider to introduce semantic priors into the system.

In parallel, the depth branch operates on a stacked input that includes RGB images, sparse depth measurements, and a binary validity mask. This combined input is passed through a U-Net architecture that produces two outputs: a low-level dense depth prediction and a rich set of spatial feature maps. These depth features capture both global structure and fine detail, which are essential for high-quality completion.

The cross-attention module aligns these two streams. The instance mask acts as a spatial guide, allowing the model to reweight the depth features using attention mechanisms that prioritize object-centric regions. This produces an enhanced feature map that highlights areas likely to contain meaningful geometric structure.

The fusion block combines the attention-enhanced features with the original depth features. This fused representation is then passed to the prediction head, which includes a channel attention mechanism followed by a convolutional decoder that outputs the final dense depth map.

\begin{figure}[!h] 
    \centering
    \includegraphics[width=0.9\linewidth]{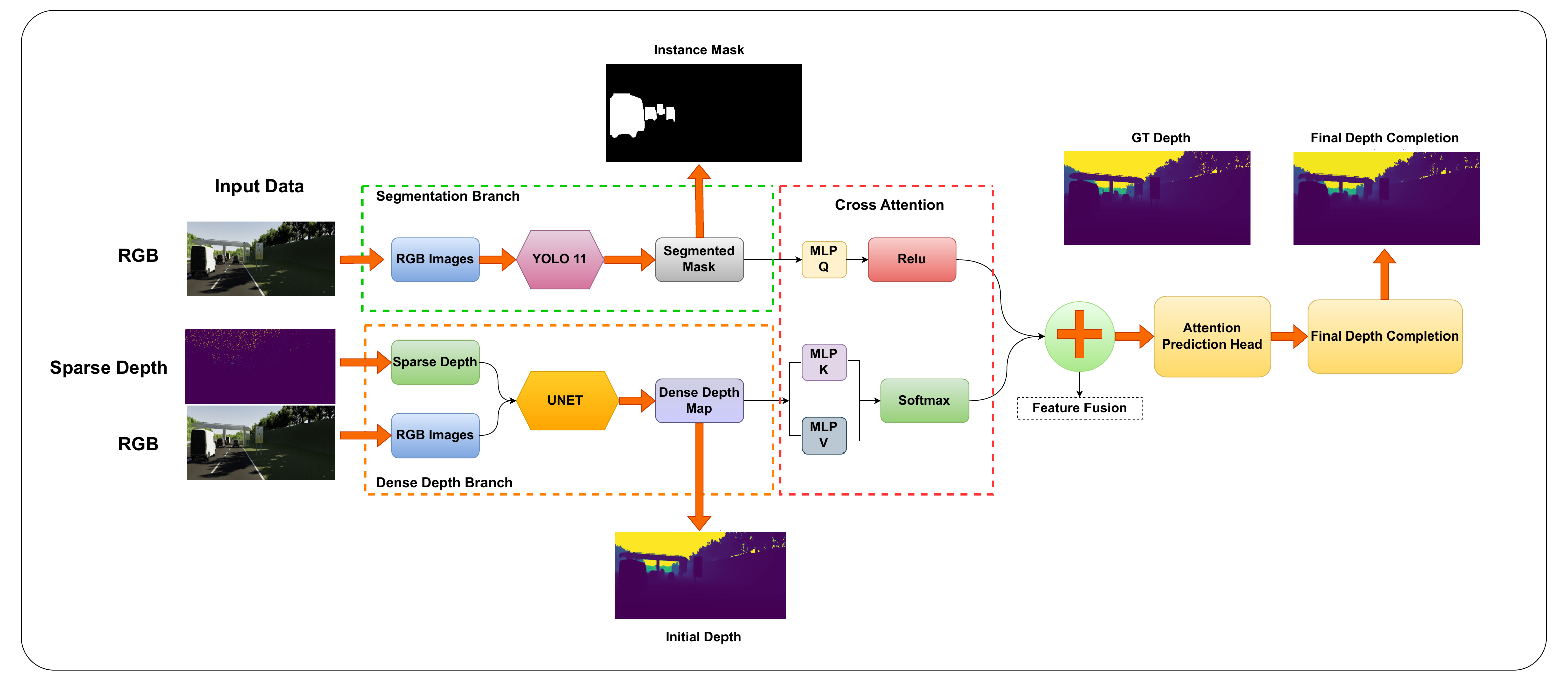}
    \caption{Block Diagram of Proposed MT-Depth Model}
    \label{fig1}
\end{figure}

During training, we supervise both the initial and final depth outputs using ground-truth depth maps. We also enforce consistency between the predicted instance mask and the ground-truth instance segmentation to improve semantic alignment. Finally, we evaluate the model's depth accuracy using standard regression metrics — Mean Absolute Error (MAE) and Root Mean Squared Error (RMSE) — over valid pixels. Figure \ref{fig1} demonstrate the architecture of our proposed MT-Depth model.

\begin{algorithm}[h]
\caption{Semantic-Guided Depth Completion with YOLO V11 and U-Net}
\label{alg:depth_completion}

{
    RGB image \texttt{I} $\in \mathbb{R}^{3 \times H \times W}$ \\
    Sparse depth map \texttt{D\_sparse} $\in \mathbb{R}^{1 \times H \times W}$ \\
    Validity mask \texttt{V} $\in \{0,1\}^{1 \times H \times W}$ \\
    Ground-truth depth \texttt{D\_gt} $\in \mathbb{R}^{1 \times H \times W}$ \\
    Ground-truth instance mask \texttt{M\_gt} $\in \{0,1\}^{1 \times H \times W}$
}

\textbf{Step 1: Instance Mask Extraction using YOLO V11} \;
Freeze the YOLO V11 model and run it on input RGB image \texttt{I} \;
Obtain $n$ instance masks and merge them into a binary mask \texttt{M\_seg} of shape $1 \times H \times W$ \;

\textbf{Step 2: Depth Feature Extraction with U-Net} \;
Concatenate \texttt{I}, \texttt{D\_sparse}, and \texttt{V} into a 5-channel input of shape $5 \times H \times W$ \;
Pass through U-Net to obtain: \;

Initial dense depth prediction \texttt{D\_init} $\in \mathbb{R}^{1 \times H \times W}$ \;
Depth features \texttt{F\_depth} $\in \mathbb{R}^{C \times H \times W}$ \;

\textbf{Step 3: Cross-Attention Fusion} \;
Downsample \texttt{M\_seg} and \texttt{F\_depth} to a smaller size (e.g., $16 \times 32$) \;
Use cross-attention where \texttt{M\_seg} is the query and \texttt{F\_depth} is the key/value \;
Obtain attention-enhanced features \texttt{F\_att} $\in \mathbb{R}^{C \times H \times W}$ after upsampling \;

\textbf{Step 4: Feature Fusion} \;
Concatenate \texttt{F\_att} and \texttt{F\_depth} along the channel axis \;
Apply $1 \times 1$ convolution, BatchNorm, and ReLU to get fused features \texttt{F\_fused} $\in \mathbb{R}^{C' \times H \times W}$ \;

\textbf{Step 5: Final Depth Prediction (Attention Head)} \;
Apply channel attention to \texttt{F\_fused} to reweight channels \;
Pass through convolutional layers to obtain final dense depth prediction: \;

\texttt{D\_final} $\in \mathbb{R}^{1 \times H \times W}$ \;

\textbf{Step 6: Evaluation} \;
Evaluate only on valid pixels where \texttt{D\_gt} > 0 \;
Compute Mean Absolute Error (MAE) and Root Mean Squared Error (RMSE) between \texttt{D\_final} and \texttt{D\_gt} \;

\textbf{Output:}\\
Initial dense depth prediction: \texttt{D\_init} $\in \mathbb{R}^{1 \times H \times W}$ \\
Final dense depth prediction: \texttt{D\_final} $\in \mathbb{R}^{1 \times H \times W}$ \\
MAE and RMSE scores for depth evaluation \\
Instance mask prediction: \texttt{M\_seg} $\in \mathbb{R}^{1 \times H \times W}$

\end{algorithm}

\section{Results and Experiment}

\subsection{Virtual KITTI 2 Dataset}

We use the Virtual KITTI 2 dataset, a photorealistic synthetic clone of real-world KITTI scenes, designed for training and evaluating vision models under controlled conditions \cite{cabon2020vkitti2, Gaidon2016vkitti}. It provides synchronized RGB images, dense depth maps, instance segmentation, and class-level semantic labels across multiple virtual scenes and weather conditions. Each frame is rendered from multiple cameras and includes perfect ground truth, making it ideal for depth completion tasks. In our experiments, we use data from the front-facing camera (Camera\_0), resizing all inputs to $256 \times 512$ and applying an 80/20 split for training and validation \ref{tab:vkitti_sample}.

\begin{table}[h]
\centering
\caption{Example of data sample structure from Virtual KITTI 2.}
\label{tab:vkitti_sample}
\begin{tabular}{|l|l|}
\hline
\textbf{Field} & \textbf{Description} \\
\hline
scene          & Scene ID (e.g., 0001, 0002) \\
condition      & Lighting/weather (e.g., clone, overcast) \\
rgb            & RGB image path ($3 \times H \times W$) \\
depth          & Dense depth map ($1 \times H \times W$) \\
instance       & Instance segmentation mask ($1 \times H \times W$) \\
class\_seg     & Class-level semantic mask  \\
\hline
\end{tabular}
\end{table}

\subsection{Experimental Setup}

All experiments were conducted on a Windows 11 system equipped with high-performance hardware. The machine was powered by an NVIDIA GeForce RTX 5080 GPU with 16 GB of VRAM and an Intel(R) Core(TM) Ultra 7 255HX processor clocked at 2.40 GHz. It included 32 GB of DDR5 RAM operating at 5600 MT/s. CUDA version 12.8 and NVIDIA driver version 572.90 were used for GPU acceleration.

The model was implemented in PyTorch version 2.8.0 and trained using the Adam optimizer with a learning rate of $1 \times 10^{-4}$. Training was run for 100 epochs with a batch size of 4, chosen to balance GPU memory usage, especially when processing high-resolution instance masks. Mixed precision training was enabled to improve computational efficiency and reduce memory load.

We used a pretrained YOLO V11 instance segmentation model to extract binary instance masks from the RGB input images. This model was kept frozen throughout training and only served as an external feature provider. The extracted masks were merged into a single-channel binary map highlighting foreground object regions. These instance masks were then used in a cross-attention module to guide depth refinement.

The depth completion model was trained on the Virtual KITTI 2.0 dataset using an 80\%/20\% train-validation split. All RGB images, sparse depth maps, and instance masks were resized to a resolution of $256 \times 512$. Sparse depth maps were simulated by randomly dropping valid depth pixels with a keep probability of 5\%. 

Final evaluation was performed using Mean Absolute Error (MAE) and Root Mean Squared Error (RMSE), computed only on valid pixels with non-zero ground-truth depth values. These metrics were used to compare both the initial depth prediction and the final refined depth output.

\subsection{Evaluation Metrics}

We evaluate the depth completion performance using two standard regression metrics: Mean Absolute Error (MAE) and Root Mean Squared Error (RMSE), computed over valid pixels where ground-truth depth is available.

Let $\hat{\mathbf{D}}$ be the predicted dense depth map and $\mathbf{D}_{\text{gt}}$ be the ground-truth depth map, both of shape $(H \times W)$. Let $\Omega$ be the set of valid pixel indices where $\mathbf{D}_{\text{gt}}(i,j) > 0$.

The MAE and RMSE are defined as follows:

\begin{equation}
\text{MAE} = \frac{1}{|\Omega|} \sum_{(i,j) \in \Omega} \left| \hat{\mathbf{D}}(i,j) - \mathbf{D}_{\text{gt}}(i,j) \right|
\end{equation}

\begin{equation}
\text{RMSE} = \sqrt{ \frac{1}{|\Omega|} \sum_{(i,j) \in \Omega} \left( \hat{\mathbf{D}}(i,j) - \mathbf{D}_{\text{gt}}(i,j) \right)^2 }
\end{equation}

\subsection{Parameter Settings}

To ensure consistent evaluation and reproducibility, we define all model, training, and dataset-related parameters used in our experiments. The input resolution is fixed to $256 \times 512$ for all modules, and we use a batch size of 4 due to memory constraints when using high-resolution instance masks. The depth backbone is implemented using a U-Net architecture with five convolutional stages and channel widths progressively increasing from 8 to 128. The pretrained YOLO V11 segmentation model is used in a frozen state and is not updated during training.

Sparse depth inputs are simulated by randomly dropping valid pixels with a keep probability of 5\%, producing inputs that resemble real-world LiDAR sparsity. For the attention module, we use a cross-attention embedding dimension of 32, and the final fusion head produces a 128-channel representation before regressing the final depth map. Loss balancing weights are selected empirically based on validation performance. Table \ref{tab:params} shows the parameters we have used in our experiment.

\begin{table}[h]
\centering
\caption{Parameter settings used in all experiments.}
\label{tab:params}
\begin{tabular}{|l|c|l|}
\hline
\textbf{Component} & \textbf{Value} & \textbf{Description} \\
\hline
Input resolution & $256 \times 512$ & Resized input image and depth map size \\
Batch size & 4 & Number of samples per training batch \\
YOLO model & V11-seg (frozen) & Pretrained instance segmentation model \\
Sparse depth keep prob. & 0.05 & Probability of keeping a valid depth pixel \\
Depth backbone & U-Net & 5-level encoder-decoder with skip connections \\
U-Net channels & (8, 16, 32, 64, 128) & Channels per block in encoder \\
Cross-attention dim & 32 & Channel dimension in attention module \\
Fusion output channels & 128 & Feature size before final depth head \\
Learning rate & $1 \times 10^{-4}$ & Optimizer learning rate (Adam) \\
Epochs & 100 & Number of full training epochs \\
Loss weights & $\lambda_{\text{init}}=0.5$, $\lambda_{\text{obj}}=3.0$, $\lambda_{\text{seg}}=1.0$ & Balancing weights for multi-task loss \\
\hline
\end{tabular}
\end{table}


\subsection{Quantitative Analysis}

We compare the performance of our proposed model with two baselines: the original U-Net depth completion model and SemSegDepth, a segmentation-guided approach reported in prior work. We evaluate each method using two standard error metrics — Mean Absolute Error (MAE) and Root Mean Squared Error (RMSE) — over valid pixels in the ground-truth depth maps. The results are summarized in Table~\ref{tab:quantitative_metrics}.

\begin{table}[h]
\centering
\caption{Quantitative comparison of depth completion models on the validation set. Lower values indicate better performance.}
\label{tab:quantitative_metrics}
\begin{tabular}{|l|c|c|}
\hline
\textbf{Method} & \textbf{MAE} $\downarrow$ & \textbf{RMSE} $\downarrow$ \\
\hline
SemSegDepth \cite{Lagos_2022} & - & 458.2 \\
UNet        & 61.1 & 1416.7 \\
\textbf{Ours} & \textbf{65.6} & \textbf{389.5} \\
\hline
\end{tabular}
\end{table}

The U-Net-only model produces low MAE but suffers from very high RMSE, indicating large errors in specific regions despite reasonable overall accuracy. In contrast, our model achieves the lowest RMSE (389.5), showing that incorporating semantic instance guidance helps suppress large prediction errors, especially around object boundaries and occlusions. 

Although our model's MAE (65.6) is slightly higher than U-Net (61.1), it is significantly more stable and robust across scenes, leading to much lower RMSE. Compared to SemSegDepth, our approach improves RMSE by nearly 70 units, while requiring no semantic class labels — relying only on binary instance masks extracted by YOLO V11. This confirms the benefit of object-aware fusion in depth prediction, even when using coarse segmentation priors.

\subsection{Qualitative Analysis}

\begin{figure}[h] 
    \centering
    \includegraphics[width=0.9\linewidth]{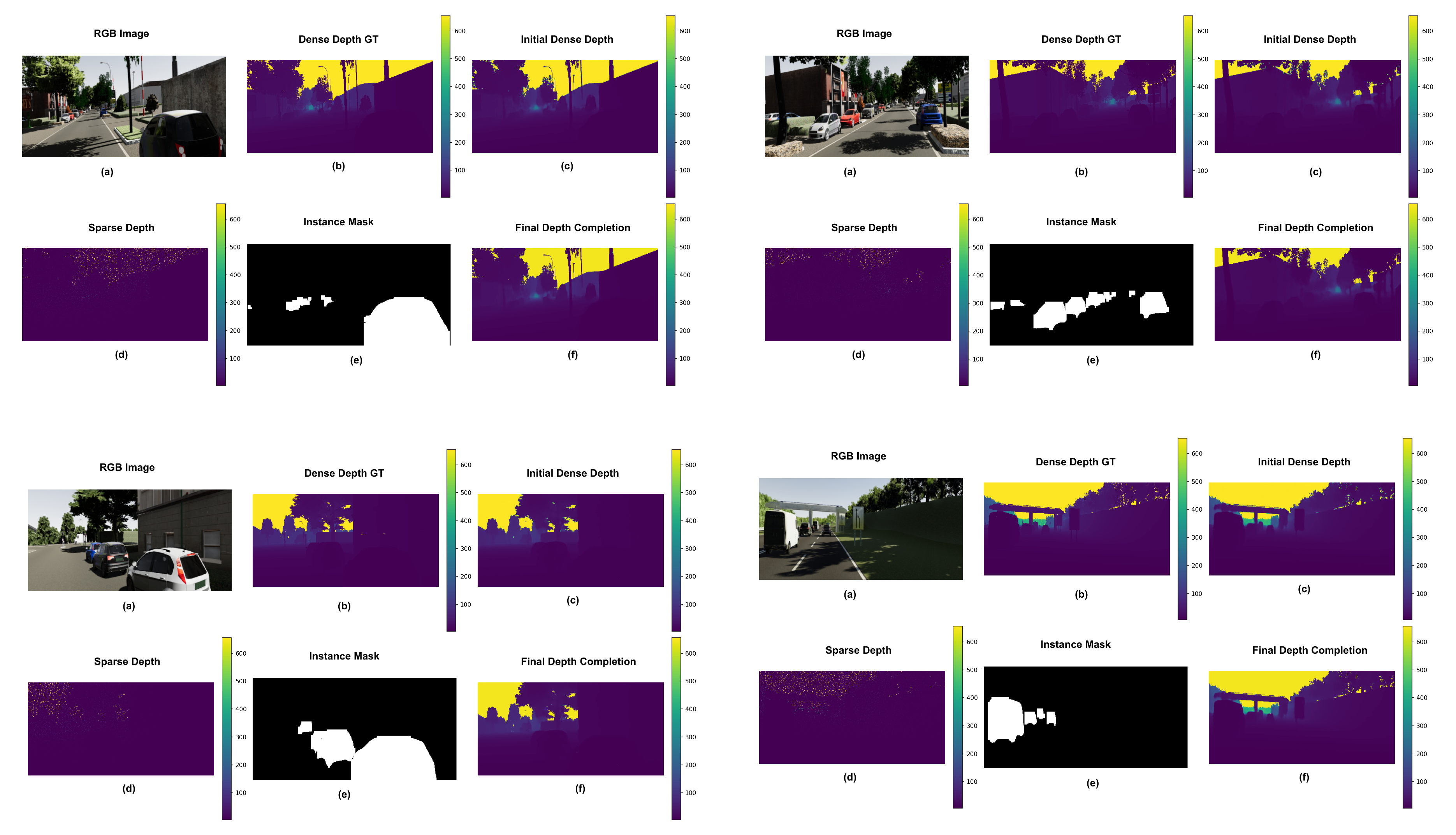}
    \caption{Qualitative results of our proposed model. (a) RGB input, (b) ground-truth depth, (c) initial prediction, (d) sparse depth, (e) instance mask, and (f) final depth completion.}
    \label{fig2}
\end{figure}

In Figure~\ref{fig2}, we visualize the full depth completion pipeline of our proposed model across multiple scenes from the Virtual KITTI 2 dataset. Each row displays the RGB input, sparse depth input, dense ground-truth depth, initial dense prediction from the U-Net backbone, instance mask extracted from the frozen YOLO V11 segmentation branch, and the final refined depth prediction. This step-by-step breakdown provides insight into how the network evolves from raw, sparse measurements to complete, high-quality depth maps through instance-aware refinement.

Notably, the final depth maps exhibit clear improvements over the initial predictions, particularly around object boundaries and occlusion regions. Foreground structures—such as vehicles, poles, and pedestrians—are more accurately reconstructed in terms of both shape and depth value. The instance masks guide the attention of the network toward semantically meaningful regions, enabling better structural alignment and continuity. In contrast to purely geometric or interpolation-based methods, our model adapts its predictions according to object-level semantics while preserving background smoothness. These qualitative results highlight the effectiveness of using instance segmentation as a spatial prior to enhance the depth completion process in cluttered, real-world environments.

\section{Conclusion}

In this paper, we introduced an instance-aware depth completion framework that leverages object-centric information to improve the accuracy and consistency of depth predictions. By integrating binary instance masks extracted from a frozen YOLO V11 segmentation model, our method guides the depth refinement process through a cross-attention mechanism that highlights foreground objects and preserves structural boundaries. The use of instance-level priors enhances the model’s ability to recover depth in challenging regions such as occlusions, thin objects, and complex scenes. Through extensive experiments on the Virtual KITTI 2 dataset, we demonstrated that our approach achieves significantly lower RMSE compared to both baseline U-Net and prior semantic-guided models, without the need for dense semantic labels. Both qualitative and quantitative results confirm the effectiveness of our instance-guided fusion strategy in producing high-quality depth maps.

\textbf{Future Work:} While our current model achieves promising results, several directions remain open for further exploration. A natural extension would involve replacing the U-Net backbone with a large-scale pretrained foundation model such as Omni-DC or Depth Anything, which have shown strong generalization across diverse depth tasks. Incorporating these models could improve the base predictions and allow better transferability to real-world scenarios. Additionally, we plan to extend our framework to handle real-world datasets and investigate the integration of temporal consistency for video-based depth completion. Incorporating panoptic segmentation or uncertainty estimation may also further refine instance-guided depth refinement and enhance the reliability of predictions in safety-critical applications such as autonomous navigation and robotic interaction.

\end{document}